\documentclass[runningheads]{llncs}
\usepackage{times}
\usepackage{latexsym}
\usepackage{graphicx}
\usepackage{url}
\usepackage{multirow}
\usepackage{amsmath}
\usepackage{graphics}
\usepackage{tabularx}
\usepackage{booktabs,amsfonts,dcolumn}
\usepackage{url}

\begin{document}

\title{Extracting N-ary Cross-sentence Relations using Constrained Subsequence Kernel}
%
%
\author{Sachin Pawar\inst{1,2} \and Pushpak Bhattacharyya\inst{2} \and Girish K. Palshikar\inst{1}}
\authorrunning{Pawar et al.}
%
\institute{TCS Research \& Innovation, Pune, India-411013 \and Indian Institute of Technology Bombay, Mumbai, India-400076\\
{\tt sachin7.p@tcs.com, pb@cse.iitb.ac.in, gk.palshikar@tcs.com}}
\maketitle              
\begin{abstract}
Most of the past work in relation extraction deals with relations occurring within a sentence and having only two entity arguments. 
We propose a new formulation of the relation extraction task where the relations are more general than intra-sentence relations in the sense that they may span multiple sentences and may have more than two arguments. Moreover, the relations are more specific than corpus-level relations in the sense that their scope is limited only within a document and not valid globally throughout the corpus. 
We propose a novel sequence representation to characterize instances of such relations. 
We then explore various classifiers whose features are derived from this sequence representation. For SVM classifier, we design a Constrained Subsequence Kernel which is a variant of Generalized Subsequence Kernel. We evaluate our approach on three datasets across two domains: biomedical and general domain.
\end{abstract}

\section{Introduction}
\begin{figure}\center
\includegraphics[width=\columnwidth,height=0.6\columnwidth]{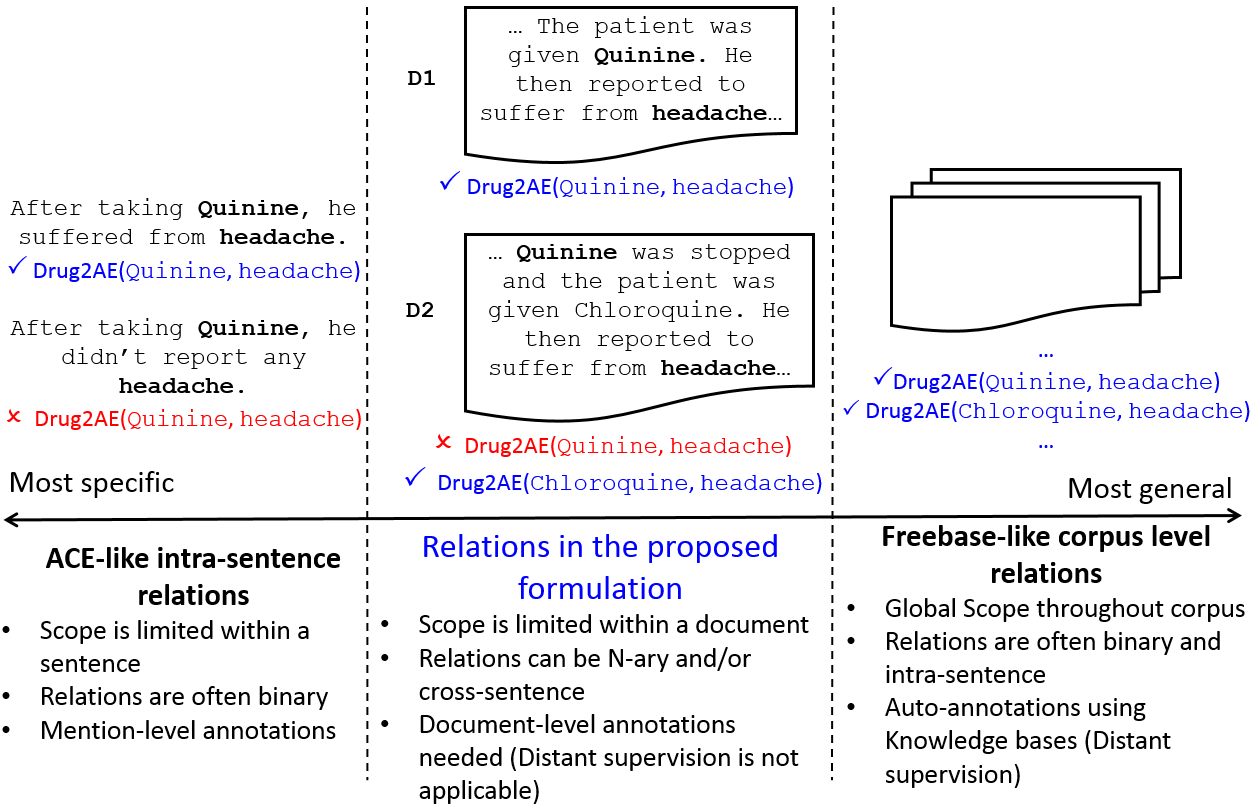}
\caption{\label{figRelationsSpectrum}Spectrum of relation extraction formulations.}
\end{figure}

The task of traditional relation extraction (RE) deals with identifying whether any pre-defined {\em semantic relation} holds between a pair of {\em entity mentions} in a given sentence~\cite{zhou2005,miwa2016end,pawar2017relation}.
In this paper, we propose a new formulation of the traditional relation extraction task, where the relations are more general than ACE-like intra-sentence relations~\cite{doddington2004automatic} in the sense that they may span multiple sentences and may have more than two entity arguments (N-ary). In addition, the relations are more specific than Freebase-like relations~\cite{bollacker2008freebase} in the sense that their scope is limited only within a document and not valid globally throughout the corpus. Hence, distant supervision based approaches~\cite{mintz2009distant,riedel2010modeling} involving corpus-level relations will not be applicable. 
E.g., {\small\sf\em Drug2AE} is such a relation between a {\small\sf\em Drug} and an {\small\sf\em AdverseEvent} it causes. Scope of the {\small\sf\em Drug2AE} relation is limited only within a particular document (case report of a patient), it may not be valid in another document (case report of another patient) even if it contains mentions of the exactly same {\small\sf\em Drug} and {\small\sf\em AdverseEvent}.
Hence, our proposed formulation is useful to retrieve only the relevant documents which actually {\em express} a relation among the entities; and not just {\em contain} the entities. Figure~\ref{figRelationsSpectrum} shows where our proposed formulation of relation extraction lies within the spectrum of traditional formulations.

In the proposed formulation, each relation type is associated with a {\em signature} which is a sequence of entity types of its arguments. 
We define a {\em candidate relation instance} to be a tuple of a document name followed by an ordered sequence of entities which are type compatible with the relation's signature. If the relation holds for the argument entities within the document, then the candidate relation instance is referred as a {\em relation instance} (Example in Table~\ref{example}).
\begin{table}[t]\small
\caption{\label{example}Example of the {\small\sf\em Succession} relation and its instances}
\begin{tabular}{p{\linewidth}}
\textbf{Relation type:} {\small\sf\em Succession}; \textbf{Signature:} ({\small\sf\em ORG}, {\small\sf\em POST}, {\small\sf\em PER}, {\small\sf\em PER}); \textbf{Meaning:} {\small\sf\em Succession}$(D, org, post, per_1, per_2)$ holds if the document $D$ reports that in the organization $org$, the person $per_1$ is succeeded by $per_2$ for the post $post$\\
\hline
Following relation instances are {\em similar} and are part of a single {\em group} (see Table~\ref{mucNews} for the corresponding document {\em news\_1.txt}):\\
$\langle${\em news\_1.txt}, \texttt{AB Volvo}, \texttt{chairman}, \texttt{Pehr G. Gyllenhammar}, \texttt{Bert-Olof Svanholm} $\rangle$\\
$\langle$ {\em news\_1.txt}, \texttt{Volvo}, \texttt{chairman}, \texttt{Pehr G. Gyllenhammar},  \texttt{Bert-Olof Svanholm} $\rangle$\\
$\langle$ {\em news\_1.txt}, \texttt{AB Volvo}, \texttt{chairman}, \texttt{Pehr G. Gyllenhammar},  \texttt{Mr. Svanholm} $\rangle$\\
$\langle$ {\em news\_1.txt}, \texttt{Volvo}, \texttt{chairman}, \texttt{Pehr G. Gyllenhammar},  \texttt{Mr. Svanholm} $\rangle$\\
\hline
\end{tabular}
\end{table}
In this new formulation, a new challenge emerges: each entity argument of a candidate relation instance may be mentioned multiple times throughout the document either as it is or in the form of its {\em aliases}; e.g. {\small\tt AB Volvo}, {\small\tt Volvo}; or {\small\tt Bert-Olof Svanholm}, {\small\tt Mr. Svanholm}. 
Two candidate relation instances are said to be {\em similar} if their corresponding argument entities are identical or aliases of each other. 
Thus, the candidate relation instances in a document can be partitioned into {\em groups} of similar instances. 
Thus, all the relation instances in Table~\ref{example} are similar and therefore part of the same group. 

Our task in this paper is to identify which of the candidate relation instances {\em truly} represent a relation type of interest. 
To characterize any candidate relation instance, we propose a novel {\em sequence representation}. 
This representation is designed in such a way that it will be same for all {\em similar} candidate relation instances in a {\em group}.
We explore various classifiers whose features are derived from this sequence representation.
We also propose a new kernel function ``Constrained Subsequence Kernel'' (CSK) which 
is designed to compute number of common subsequences of interest between any two sequence representations.
Specific contributions of this work are: 1) a new formulation of the N-ary and cross-sentence relation extraction task, 2) a novel sequence representation for characterizing the candidate relation instances, and 3) a specifically designed subsequence kernel.

\section{Problem Definition}
\begin{table}[t]\small
\caption{\label{exAnnotations} Example of annotations $\langle D_1, LE_1, LR_1\rangle$}
\begin{tabular}{p{\linewidth}}
\hline
\textbf{Target Relation:} {\small\sf\em Interact}; \textbf{Signature:} ({\small\sf\em Drug}, {\small\sf\em Gene}, {\small\sf\em Mutation}); {\small\sf\em Interact} $(D_i,d,g,m)$ holds if the document $D_i$ mentions that the drug $d$ treats the mutation $m$ in the gene $g$\\
\hline
{\bf Entity annotations $LE_1$ for document $D_1$:}\\
$\langle${\small\tt gefitinib}:{\small\sf\em Drug}$\rangle$,$\langle${\small\tt erlotinib}:{\small\sf\em Drug}$\rangle$,$\langle${\small\tt EGFR}:{\small\sf\em Gene}$\rangle$,
$\langle${\small\tt T790M}:{\small\sf\em Mutation}$\rangle$,$\langle${\small\tt A750P}:{\small\sf\em Mutation}$\rangle$, $\langle${\small\tt L858R}:{\small\sf\em Mutation}$\rangle$\\
\hline
{\bf Relation annotations $LR_1$ for document $D_1$:}\\
$\langle${\small\tt erlotinib},{\small\tt EGFR},{\small\tt L858R}$\rangle$,$\langle${\small\tt erlotinib},{\small\tt EGFR},{\small\tt T790M}$\rangle$\\
\hline
\end{tabular}
\end{table}
\noindent \textbf{Inputs:} Target relation type $R$, its signature, a set of test documents $\{\langle D'_1, LE'_1\rangle ,$ $\cdots \langle D'_L, LE'_L\rangle\}$ where $LE'_i$ is the list of entities in document $D'_i$ along with their types

\noindent \textbf{Output:} For each test document $D'_i$, all the candidate relation instances for $R$ are classified to check whether they are {\em true} relation instances to produce a list $LR'_i$ of such instances for $R$.

\noindent \textbf{Training regime:} To learn relation extraction model for $R$, we need a set of annotated training documents $\{\langle D_1, LE_1, LR_1\rangle ,\cdots$ $\langle D_K, LE_K, LR_K\rangle\}$, where $LE_i$ is the list of entities in the document $D_i$ along with their types and $LR_i$ is the list of relation instances of type $R$ which hold within $D_i$. It is enough to annotate any one representative relation instance from each {\em group} of similar instances. We refer to this annotation scheme as {\em RIGD} (at least one \textbf{R}elation \textbf{I}nstance per \textbf{G}roup per \textbf{D}ocument) level as against the traditional mention-level annotation (Table~\ref{exAnnotations}).

\noindent \textbf{Scope and Assumptions:} Multiple target relation types are handled by providing separate annotations ($LR_i$'s) and learning separate extraction models for each relation type. Identification of entities and their types is out of scope of this paper. For each test document $D'_i$, we assume availability of the $LE'_i$'s. We do not assume availability of alias information. For identifying aliases, we use high-precision domain-specific rules; details of these rules are included in the Appendix.

\noindent \textbf{Evaluation:} We evaluate at {\em RIGD}-level where a true positive is counted for each {\em group} of predicted relation instances, if any of the relation instance from the group is listed in gold-annotations ($LR_i$). 
A false positive is counted for each group of predicted relation instances, if none of the relation instances from the group are listed in gold-annotations. Also, a false negative is counted for each group of relation instances listed in gold-annotations, if none of the relation instances in the group appear as the predicted relation instances. 

\section{Proposed Approach}
\begin{table}[t]\small\center
\caption{\label{mucNews} Example of a news article {\small\sf\em news\_1.txt} where entities of interest are highlighted}
\begin{tabular}{p{\linewidth}}
\hline
\texttt{
An extraordinary shareholders meeting of \textbf{AB} \textbf{Volvo} in Gothenburg, Sweden, elected \textbf{Bert-Olof} \textbf{Svanholm} \textbf{chairman} of the Swedish automotive group, in line with an earlier proposal. \textbf{Mr.} \textbf{Svanholm} is \textbf{president} of \textbf{ABB Asea Brown Boveri Ltd.}, an engineering concern jointly owned by \textbf{Asea AB of Sweden} and \textbf{BBC Brown Boveri AG} of Switzerland. \textbf{He} succeeds \textbf{Pehr} \textbf{G.} \textbf{Gyllenhammar},who resigned in December after the collapse of a plan to merge \textbf{Volvo}'s vehicle operations with those of French partner \textbf{Renault SA}.
}\\
\hline
\end{tabular}
\end{table}
\begin{table}[t]\small\center
\caption{\label{seqRepresentation} Examples of sequence representations. We use ; for separating tokens in a sequence}
\begin{tabular}{p{\linewidth}}
\hline
\textbf{Sequence representation} of $T_1$=$\langle D$={\small\sf\em news\_1.txt}, $E_1$={\small\tt AB Volvo}, $E_2$={\small\tt chairman}, $E_3$={\small\tt Pehr G. Gyllenhammar}, $E_4$={\small\tt Bert-Olof Svanholm}$\rangle$:\\
\hline
\texttt{extraordinary}; \texttt{shareholders}; \texttt{meeting}; \texttt{of}; $E_1$; \texttt{in}; \texttt{gothenburg}; \texttt{sweden}; \texttt{elected}; $E_4$; $E_2$; \texttt{of}; \texttt{Swedish}; \texttt{automotive}; \texttt{group}; \texttt{in}; \texttt{line}; \texttt{with}; \texttt{earlier}; \texttt{proposal}; {\small\sf\em SB}; $E_4$; $OE_{POST}$; \texttt{of}; $OE_{ORG}$; \texttt{engineering}; \texttt{concern}; \texttt{jointly}; \texttt{owned}; \texttt{by}; $OE_{ORG}$; $OE_{ORG}$; \texttt{of}; \texttt{switzerland}; {\small\sf\em SB}; $E_4$; \texttt{succeeds}; $E_3$; \texttt{resigned}; \texttt{in}; \texttt{december}; \texttt{after}; \texttt{collapse}; \texttt{of}; \texttt{plan}; \texttt{to}; \texttt{merge}; $E_1$; \texttt{vehicle}; \texttt{operations}; \texttt{with}; \texttt{of}; \texttt{french}; \texttt{partner}; $OE_{ORG}$\\
\hline
\hline
\textbf{Sequence representation} of $T_2$=$\langle D$={\small\sf\em news\_1.txt}, $E_1$={\small\tt ABB Asea Brown Boveri Ltd.}, $E_2$={\small\tt president}, $E_3$={\small\tt Bert-Olof Svanholm}, $E_4$={\small\tt Pehr G. Gyllenhammar}$\rangle$:\\
\hline
\texttt{extraordinary}; \texttt{shareholders}; \texttt{meeting}; \texttt{of}; $OE_{ORG}$; \texttt{in}; \texttt{gothenburg}; \texttt{sweden}; \texttt{elected}; $E_3$; $OE_{POST}$; \texttt{of}; \texttt{swedish}; \texttt{automotive}; \texttt{group}; \texttt{in}; \texttt{line}; \texttt{with}; \texttt{earlier}; \texttt{proposal}; {\small\sf\em SB}; $E_3$; $E_2$; \texttt{of}; $E_1$; \texttt{engineering}; \texttt{concern}; \texttt{jointly}; \texttt{owned}; \texttt{by}; $OE_{ORG}$; $OE_{ORG}$; \texttt{of}; \texttt{switzerland}; {\small\sf\em SB}; $E_3$; \texttt{succeeds}; $E_4$; \texttt{resigned}; \texttt{in}; \texttt{december}; \texttt{after}; \texttt{collapse}; \texttt{of}; \texttt{plan}; \texttt{to}; \texttt{merge}; $OE_{ORG}$; \texttt{vehicle}; \texttt{operations}; \texttt{with}; \texttt{of}; \texttt{french}; \texttt{partner}; $OE_{ORG}$\\
\hline
\end{tabular}
\end{table}


We propose a novel \textbf{sequence representation} for a candidate relation instance, which captures its important characteristics. 
An N-ary candidate relation instance is a $(N+1)$ tuple containing document name $D$ and $N$ entity arguments--$E_1, E_2,\cdots E_N$. 
Here, $E_1,\cdots E_N$ may not be of $N$ {\em distinct} entity types; $E_i$ just represents the $i^{th}$ entity argument of the relation. 
Sequence of entity types of these $N$ entity arguments ($ET_1, ET_2, \cdots ET_N$) is the {\em signature} of the relation type. 
E.g., the relation {\small\sf\em Succession} (Table~\ref{example}) holds between 4 entity arguments of entity types: $ET_1$= {\small\sf\em ORG}, $ET_2$= {\small\sf\em POST} and $ET_3$=$ET_4$= {\small\sf\em PER}. An example instance of this relation is represented as the following 5-tuple (Table~\ref{mucNews}): $T_1$= $\langle D$= {\small\sf\em news\_1.txt}, $E_1$= {\small\tt AB Volvo}, $E_2$= {\small\tt chairman}, $E_3$= {\small\tt Pehr G. Gyllenhammar}, $E_4$= {\small\tt Bert-Olof Svanholm}$\rangle$.
There are other candidate relation instances for which the relation does not hold but they just conform to its signature; e.g., $T_2$= $\langle D$= {\small\sf\em news\_1.txt}, $E_1$= {\small\tt ABB Asea Brown Boveri Ltd.}, $E_2$= {\small\tt president}, $E_3$= {\small\tt Bert-Olof Svanholm}, $E_4$= {\small\tt Pehr G. Gyllenhammar}$\rangle$.
Hence, $T_1$ is a positive instance for the relation type {\small\sf\em Succession} whereas $T_2$ is a negative instance. 


\noindent \textbf{Span and Minimal Span:} We define {\em span} of a candidate relation instance $T$ as the sequence of sentences in the document $D$ covering all the mentions of its argument entities (including aliases). The sequence starts with the earliest mention of any entity argument (or its alias) involved in $T$ and stretches up to the latest mention. For each argument pair of $T$, minimum number of sentences separating corresponding entity mentions is computed. 
We define {\em minimal span} of a candidate relation instance as the maximum of minimum number of separating sentences across all argument pairs. We use 
{\em minimal span} to filter out candidate relation instances having values more than some threshold. 
E.g., $T_1$ has span of 3 sentences and the minimal span of 2 (Table~\ref{mucNews}) corresponding to the argument pair of $E_2$ and $E_3$ which are 2 sentences apart and this is the maximum separation across all argument pairs.

\subsection{Constructing Sequence Representations}
We propose to characterize any candidate relation instance $T$ of relation type $R$ in the form of \textbf{a sequence of {\em tokens} of certain types}:\\
$\bullet$ $E_i$: Mentions of the $i^{th}$ entity argument (and its aliases) of $T$ within the span of $T$\\
$\bullet$ {\small\sf\em SB}: Sentence boundaries of the sentences in the {\em span} of $T$\\
$\bullet$ $OE_{ET_j}$: Mentions of {\em other} entities than the argument entities (and their aliases) which occur within the span of $T$ and which are of type $ET_j$. Tokens of this type encode important discourse information by capturing mentions of other entities of type $ET_j$.\\
$\bullet$ Words: All the words 
(excluding stop words)
occurring within the span of the instance $T$.

In order to construct the sequence representation of a candidate relation instance, these tokens are arranged sequentially from the beginning of the {\em span} till the end. The tokens are arranged in the same order as they occur in the document. In other words, these tokens are place-holders in the sequence representation, for each important piece of information.  
Table~\ref{seqRepresentation} shows sequence representations for our example instances $T_1$ and $T_2$. 
It can be observed that for $T_1$, all the mentions of the entity argument {\small\tt Bert-Olof Svanholm} including its aliases ({\small\tt Mr. Svanholm} and {\small\tt He})\footnote{In principle, coreferences of entity arguments can be used instead of just aliases when we add tokens of the form $E_i$ and $OE_{ET_j}$ in the sequence representation. But coreference resolution is itself a difficult problem and is not accurate enough in practice for Biomedical domain documents. Hence, we use coreferences only for general domain dataset.} are captured using the token $E_4$. Also, {\small\tt president} is represented using the token $OE_{POST}$, which is an entity of type {\small\sf\em POST} but is not an argument entity of $T_1$ .

\noindent \textbf{Generalizing the sequence representation:} 
We create clusters of the frequently occurring words in a domain. By considering cosine similarity among the word embeddings, we apply hierarchical clustering with complete linkage. 
E.g., in Biomedical domain, following is an example word cluster \{{\small\tt radiotherapy}, {\small\tt chemotherapy}, {\small\tt adjuvant}, {\small\tt immunotherapy}\}. 
These word clusters are used to generalize the sequence representation by replacing a single word token with a set of 2 tokens containing word itself along with its cluster ID. Hence, the sequence representation for $T_1$ would now be:
\small
\{c12, \texttt{extraordinary}\}; \{c43, \texttt{shareholders}\}; \{c47, \texttt{meeting}\}; \texttt{of}; $E_1$; \texttt{in};$\cdots$
\normalsize


\subsection{Constrained Subsequence Kernel (CSK)}
Generalized Subsequence Kernel (GSK) proposed by Mooney and Bunescu~\cite{mooney2005} computes the number of common subsequences of length $n$ shared by two {\em generalized} sequences, in polynomial time. The sequences which share more such common subsequences, get a higher similarity score. 
Moreover, the subsequences are weighted by their sparseness in the original sequences, i.e. subsequences which are not contiguous and spread over a greater length in the sequences will get lower weights (inversely proportional to length of their spread in those sequences). In a {\em generalized} sequence, any token in a sequence can be generalized to a set of values; 
e.g., we use word cluster IDs as the generalizations for actual words.

\begin{table}[t]\small
\caption{\label{exSubsequences} A few example subsequences of sequence representations of $T_1$ and $T_2$ (Table~\ref{seqRepresentation})}
\begin{tabular}{p{\columnwidth}}
\hline
\textbf{Subsequences not satisfying Constraint 1:}
(\texttt{meeting}; $OE_{ORG}$; \texttt{elected}; $E_3$), 
($E_4$; $OE_{POST}$; \texttt{of})\\
\hline
\textbf{Subsequences not satisfying Constraint 2:}
($E_1$; $E_2$), 
($E_3$; $E_3$), ($E_2$; $E_4$)\\
\hline
\textbf{Subsequences satisfying both the constraints:}
(\texttt{meeting}; $E_1$; \texttt{elected}; $E_4$), 
($E_4$; \texttt{succeeds}; $E_3$), 
(\texttt{elected}; $E_4$; $E_2$; \texttt{of}; \texttt{group}), 
($E_2$; \texttt{of}; $E_1$; {\small\sf\em SB}; $E_3$)\\
\hline
\end{tabular}
\end{table}
We propose a variant of the generalized subsequence kernel, namely {\em Constrained Subsequence Kernel} (CSK). As the name suggests, CSK differs from the original kernel GSK by constraining the subsequences to consider.
Our goal is to design a kernel function such that it will compute a high similarity score among the two candidate relation instances if both of them are {\em true} relation instances. The similarity should be lower if one of them is a true relation instance and other is not. Hence, the intuition is that the common subsequences (to consider during kernel computation) should contain at least two distinct tokens of the type $E_i$. Because presence of at least two of these tokens in a subsequence, ensures that the subsequence captures interaction among at least two entity arguments. Also, the common subsequences are constrained to have length of at least 3. This ensures that any common subsequence will contain at least one token other than the two tokens corresponding to entity arguments.
Thus, the following 2 constraints are considered (see Table~\ref{exSubsequences}).

\begin{center}
\textbf{Constraint 1:} A subsequence should contain \textbf{at least two distinct tokens from} $E_1, E_2,\cdots, E_N$\\
\textbf{Constraint 2:} A subsequence should contain \textbf{at least three tokens}
\end{center}

\subsection{Formal Definition of CSK}
\begin{table}[b]\small\center
\caption{\label{auxFunctions} Additional auxiliary functions defined analogous to $K'_n$ and $K''_n$}
\begin{tabular}{p{0.18\columnwidth}p{0.7\columnwidth}}
\hline
\textbf{Auxiliary Functions} & \textbf{For counting common subsequences which...} \\
\hline
$aK'_n,aK''_n$ & contain token $a$\\
\hline
$bK'_n,bK''_n$ & contain token $b$\\
\hline
$abK'_n,abK''_n$ & contain both the tokens $a$ and $b$\\
\hline
\end{tabular}
\normalsize
\end{table}
Let $CSK(s,t,n,\lambda,a,b)$ be the constrained subsequence kernel which computes number of $\lambda$-weighted common subsequences of length $n$ shared by the sequences $s$ and $t$ such that each of these common subsequences contains particular tokens $a$ and $b$. Here, $a$ and $b$ are considered to incorporate the \textit{Constraint 1}; and the \textit{Constraint 2} is trivially satisfied by considering $CSK$ with $n\ge 3$. Also, $\lambda$ is a number between 0 and 1. Each common subsequence is weighted by $\lambda^l$ where $l$ is the sum of lengths of the subsequence's spread in $s$ and $t$. 
Let $\Sigma_1, \Sigma_2, \Sigma_3,$ $\Sigma_4$ and $\Sigma_5$ be disjoint spaces representing various types of tokens used in the sequence representation: 
(i) $\Sigma_1=\{E_1, E_2,\cdots E_N\}$, for $N$-ary relation type, $E_i$ is a place-holder for mentions of the $i^{th}$ entity argument of the relation type, 
(ii) $\Sigma_2=\{${\small\sf\em SB}$\}$, the ``sentence break'' token, 
(iii) $\Sigma_3=\{OE_{ET_1}, \cdots OE_{ET_M}\}$, 
$M$ is the no. of distinct entity types in the relation's signature, 
(iv) $\Sigma_4$=Set of words, (v) $\Sigma_5$=Set of word cluster IDs

Sequence representation for any candidate relation instance belongs to $\Sigma^{*}$, where $\Sigma = \Sigma_1 \cup \Sigma_2 \cup \Sigma_3 \cup \{\Sigma_4 \times \Sigma_5\}$. 
Each sparse subsequence of such sequence representations then belongs to $\Sigma'^{*}$ where $\Sigma' = \Sigma_1 \cup \Sigma_2 \cup \Sigma_3 \cup \Sigma_4 \cup \Sigma_5$. 
We design an efficient recursive formulation for computing $CSK(s,t,n,\lambda,a,b)$. 
This formulation is derived on the similar lines as that of the generalized subsequence kernel~\cite{mooney2005} and hence the same notations ($K_n(s,t)$, $K'_n(s,t)$ and $K''_n(s,t)$) are used with similar meaning. However, we introduce additional auxiliary functions (see Table~\ref{auxFunctions}) for taking the {\em Constraint 1} into consideration for $CSK$ computation. As per its definition, $K'_n(s,t)$ adds length from beginning of a common subsequence to the end of the sequences $s$ and $t$. We define $aK'_n(s,t)$ to be a similar function with the only difference that it considers only those common subsequences which contain the token $a$. Similarly, the function $aK''_n(s,t)$ is defined analogous to $K''_n(s,t)$. The other auxiliary functions are defined analogously. The detailed computation steps are depicted in the Table~\ref{CSK_Computation}. 
Here, the function $c(x,y)$ computes the number of common tokens between the sets $x$ and $y$. 
Also, as $a$ and $b$ would always be from $\{E_1, E_2,\cdots E_N\}$ and there are no generalizations defined for these tokens, the equality conditions ($x==a$ and $x==b$) will be satisfied only when singleton tokens are involved. This simplifies the computation for additional auxiliary functions. Recursive computations of $K'$ and $K''$ are same as the generalized subsequence kernel. Table~\ref{CSK_Computation} shows recursive updates for other functions (like $aK'', bK'', abK''$, $aK', bK'$ and $abK'$) by taking into consideration the first constraint. These values are computed incrementally from $i = 0$ to $n$ and finally used to compute the value of $abK_n(s,t)$ i.e. $CSK(s,t,n,\lambda,a,b)$.
\begin{table*}[!t]\center
\caption{\label{CSK_Computation} Recursive and efficient computation of Constrained Subsequence Kernel (CSK)}
\begin{tabular}{l}
\hline
Recursive computation for $abK_n(s,t)=CSK(s,t,n,\lambda,a,b)$:\\
\hline
$K'_0 (s,t) = 1$, for all $s,t$\\
$K''_i (sx,ty) = \lambda K''_i (sx,t) + \lambda^2 K'_{i-1} (s,t) c(x,y)$\\
{\bf If} $x == y$ {\bf and} $x == a$ {\bf then},\\
\hspace{4mm}$aK''_i (sx,ty) = \lambda aK{''}_i (sx,t) + \lambda^2 K'_{i-1} (s,t)$\\
\hspace{4mm}{\bf If} $i > 1$ {\bf then},\\
\hspace{8mm}$bK''_i (sx,ty) = \lambda bK''_i (sx,t) + \lambda^2 bK'_{i-1} (s,t)$\\
\hspace{8mm}$abK''_i (sx,ty) = \lambda abK''_i (sx,t) + \lambda^2 bK'_{i-1} (s,t)$\\
{\bf Else If} $x == y$ {\bf and} $x == b$ {\bf then},\\
\hspace{4mm}$bK''_i (sx,ty) = \lambda bK''_i (sx,t) + \lambda^2 K'_{i-1} (s,t)$\\
\hspace{4mm}{\bf If} $i > 1$ {\bf then},\\
\hspace{8mm}$aK''_i (sx,ty) = \lambda aK''_i (sx,t) + \lambda^2 aK'_{i-1} (s,t)$\\
\hspace{8mm}$abK''_i (sx,ty) = \lambda abK''_i (sx,t) + \lambda^2 aK'_{i-1} (s,t)$\\
{\bf Else If} $c(x,y) > 0$ {\bf then},\\
\hspace{4mm}{\bf If} $i > 1$ {\bf then},\\
\hspace{8mm}$aK''_i (sx,ty) = \lambda aK''_i (sx,t) + \lambda^2 aK'_{i-1} (s,t) c(x,y)$\\
\hspace{8mm}$bK''_i (sx,ty) = \lambda bK''_i (sx,t) + \lambda^2 bK'_{i-1} (s,t) c(x,y)$\\
\hspace{8mm}$abK''_i (sx,ty) = \lambda abK''_i (sx,t) + \lambda^2 abK'_{i-1} (s,t) c(x,y)$\\
{\bf Else}\\
\hspace{4mm}$aK''_i (sx,ty) = \lambda aK''_i (sx,t)${\bf ;}\hspace{1mm}$bK''_i (sx,ty) = \lambda bK''_i (sx,t)$\\
\hspace{4mm}$abK''_i (sx,ty) = \lambda abK''_i (sx,t)$\\
\\
$K'_i (sx,t) = \lambda K'_i (s,t) + K''_i (sx,t)${\bf ;}\hspace{1mm}$aK'_i (sx,t) = \lambda aK'_i (s,t) + aK''_i (sx,t)$\\
$bK'_i (sx,t) = \lambda bK'_i (s,t) + bK''_i (sx,t)${\bf ;}\hspace{1mm}$abK'_i (sx,t) = \lambda abK'_i (s,t) + abK''_i (sx,t)$\\
\\
$Sum := 0$\\
{\bf For} $j=1$ to $|t|$\\
\hspace{4mm} {\bf If} $x = t[j]$ {\bf and} $x = a$ {\bf then}, $Sum := Sum + \lambda^2 bK'_{n-1} (s,t[1:j-1])$\\
\hspace{4mm} {\bf Else If} $x = t[j]$ {\bf and} $x = b$ {\bf then}, $Sum := Sum + \lambda^2 aK'_{n-1} (s,t[1:j-1])$\\
\hspace{4mm} {\bf Else If} $c(x,t[j]) > 0$ {\bf then}, $Sum := Sum + \lambda^2 abK'_{n-1} (s,t[1:j-1]) c(x, t[j])$\\
$abK_n (sx,t) = abK_n (s,t) + Sum$\\
\hline
\end{tabular}
\end{table*}

\subsection{Classifying Candidate Relation Instances}
For each training document $D_i$ (Table~\ref{exAnnotations}), initially candidate relation instances are generated using all possible combinations of entities in $LE_i$ which conform to the signature of the target relation type. Candidate relation instances having {\em minimal span} more than some threshold are filtered out as they are unlikely to be true relation instances. Out of the remaining candidate relation instances, the ones which are {\em similar} to any of the instance in $LR_i$ are treated as positive instances for a {\em binary} classifier and the remaining as negative instances. 
During testing, candidate relation instances are generated for a document $D'_i$ in a similar way by using $LE'_i$ and the signature of the target relation type.
We explored 3 classifiers whose features are derived from the proposed sequence representation, either explicitly (MaxEnt) or implicitly (SVM with CSK \& LSTM).
\begin{table*}\small\center
\caption{\label{featuresMaxent} MaxEnt features for a candidate relation instance $T$ and its sequence repr. $Seq(T)$
}
\begin{tabular}{p{0.18\linewidth}p{0.82\linewidth}}
\hline
\textbf{Feature} & \textbf{Description}\\
\hline
$TupleSpan$ & Integer-valued feature indicating the minimal span of $T$ in terms of number of sentences\\
\hline
$SentDiff_{ij}$ $SameLine_{ij}$& Integer-valued features for each pair of $E_i$ \& $E_j$ indicating minimum number of sentences separating them in $Seq(T)$; or indicating whether they occur in a single sentence in the span of $T$\\
\hline
$E_iE_jOE_{ET_k}$ & Boolean feature for each triplet of $E_i,E_j$ and $OE_{ET_k}$ which is true if $OE_{ET_k}$ occurs between $E_i$ and $E_j$ in $Seq(T)$; and $ET_k$ is entity type of either $E_i$ or $E_j$. These features capture key discourse information about mentions of other entities occurring between the mentions of argument entities of $T$.\\
\hline
$E_iE_jNoOE_{ET_k}$ & Boolean feature for each triplet of $E_i,E_j$ and $OE_{ET_k}$ which is true if no $OE_{ET_k}$ occurs between $E_i$ and $E_j$ in $Seq(T)$; and $ET_k$ is entity type of either $E_i$ or $E_j$\\
\hline
Word / Cluster & Each word occurring in $Seq(T)$ and its cluster ID are boolean features because some of these words may be key lexical cue-words for the relation type\\
\hline
\end{tabular}
\end{table*}
\normalsize

\noindent \textbf{Maximum Entropy (MaxEnt) Classifier}~\cite{berger1996maximum}:
The features 
are explicitly engineered from the sequence representation 
of any candidate relation instance (Table~\ref{featuresMaxent}). 

\noindent \textbf{LSTM}~\cite{hochreiter1997long}\textbf{-based Classifier}:
We define an embedded representation for each unique token appearing in 
sequence representations. This representation is a concatenation of two vectors. For word tokens, the first vector is initialized using pre-trained word embeddings and the second vector is set to all zeros. For other tokens (e.g. $E_i$, $OE_{ET_j}$, {\small\sf\em SB}), the first vector is set to all zeros whereas the second vector contains one-hot representation for all the distinct non-word tokens. Sequence of these tokens is then passed through an LSTM layer and the output of the final step is connected through a hidden layer to a softmax layer representing the two class labels.

\noindent \textbf{Support Vector Machines with CSK:}
Our principal approach is SVM~\cite{cortes1995support} with the CSK kernel. Let $R$ be an $N$-ary relation type and $s, t$ be the sequence representations of any two candidate relation instances. Let $CSK^{\lambda}_N$ be the overall kernel across all entity arguments of $R$. It is computed and normalized as follows:

\small
\begin{eqnarray*}
CSK^{\lambda}_N(s,t,n)=\sum_{i=1}^{N-1} \sum_{j=i+1}^N CSK(s,t,n,\lambda,E_i,E_j)\\
NCSK^{\lambda}_N(s,t,n) = \frac{CSK^{\lambda}_N(s,t,n)}{\sqrt{CSK^{\lambda}_N(s,s,n)\cdot CSK^{\lambda}_N(t,t,n)}}
\end{eqnarray*}
\normalsize
We combine kernel functions for various subsequence lengths (i.e. $n$) to get final kernel function:
\par\nobreak{\parskip0pt 
		\setlength{\abovedisplayskip}{0pt}
		\setlength{\belowdisplayskip}{4pt}
		\begin{small}
			\begin{align}
CSK^{\lambda}_{Final}(s,t) = \frac{\sum_{k=3}^{N'} 2^{N'-k}\cdot NCSK^{\lambda}_N(s,t,k)}{\sum_{k=3}^{N'} 2^{N'-k}}
			\nonumber
			\end{align}
		\end{small}
$N'$ is a parameter controlling the number of different subsequence lengths. E.g., 
if $N'=5$ then subsequences of lengths $3,4$ and $5$ are considered. 


\section{Experimental Analysis}
We evaluate our approach on 2 datasets from Biomedical domain and 1 general domain dataset. For the proposed new formulation of the N-ary cross-sentence relation extraction task, there are no readily available public datasets. We converted\footnote{Conversion and evaluation scripts for all the datasets will be made public if the paper is accepted.} annotations of some public datasets used for other tasks, to the {\em RIGD}-level annotations.
\subsection{Datasets}
\noindent\textbf{Bacteria Biotope:}
Bacteria Biotope task~\cite{deleger2016overview} was held as a part of BioNLP 2016 shared task. It has annotations for 
{\em Lives\_In} event having two entity arguments--{\small\sf\em Bacteria} and the location where it was found (either {\small\sf\em Habitat} or {\small\sf\em Geographical}). We mapped this event to a binary, cross-sentence relation {\small\sf\em Lives\_In}, 
and converted the mention-level annotations to {\em RIGD} level.
We used simple rules to identify aliases of the bacteria names. e.g. {\small\tt Salmonella Typhimurium}$\leftrightarrow${\small\tt S. Typhimurium} and {\small\tt salmonellae}$\leftrightarrow${\small\tt salmonella}. 
We used {\em train} and {\em dev} partitions of the dataset\footnote{\url{http://2016.bionlp-st.org/tasks/bb2}} to carry out 2-fold cross-validation (Table~\ref{tabBB_dataset})
, similar to VERSE~\cite{lever2016verse} which was the best performing system for this task. 
\begin{table}[t]\center\small
\caption{\label{tabBB_dataset} Mention-level and RIGD-level evaluation results for the binary {\small\sf\em Lives\_in} relation (Bacteria Biotope)}
\begin{tabular}{lccccccccccccc}
\hline
{\bf Level} & {\bf Fold} & \multicolumn{3}{c}{\bf MaxEnt} & \multicolumn{3}{c}{\bf LSTM} & \multicolumn{3}{c}{\bf SVM with CSK} & \multicolumn{3}{c}{\cite{lever2016verse}}\\
 &  & \hspace{2mm}{\bf P}\hspace{1mm} & {\bf R} & \hspace{1mm}{\bf F}\hspace{2mm} & \hspace{2mm}{\bf P}\hspace{1mm} & {\bf R} & \hspace{1mm}{\bf F}\hspace{2mm} & \hspace{2mm}{\bf P}\hspace{1mm} & {\bf R} & \hspace{1mm}{\bf F}\hspace{2mm} & \hspace{2mm}{\bf P}\hspace{1mm} & {\bf R} & \hspace{1mm}{\bf F}\\
\hline
 & train\_dev & 61.2 & 55.8 & 58.4 & 62.8 & 54.9 & 58.5 & 69.0 & 57.3 & \textbf{62.6} & 58.2 & 61.0 & 59.6 \\
Mention & dev\_train & 46.7 & 55.9 & 50.9 & 37.8 & 47.7 & 42.2 & 60.1 & 52.3 & \textbf{55.9} & 46.9 & 55.2 & 50.7 \\
\cline{2-14}
 & Average & 54.0 & 55.9 & 54.7 & 50.3 & 51.3 & 50.4 & 64.6 & 54.8 & \textbf{59.3} & 52.6 & 58.1 & 55.2 \\
\hline
\hline
 & train\_dev & 53.8 & 50.9 & 52.3 & 54.2 & 46.7 & 50.2 & 63.2 & 55.2 & \textbf{58.9} & \_ & \_ & \_\\
RIGD & dev\_train & 43.5 & 59.5 & 50.3 & 35.4 & 52.4 & 42.3 & 56.2 & 54.2 & \textbf{55.2} & \_ & \_ & \_\\
\cline{2-14}
 & Average & 48.7 & 55.2 & 51.3 & 44.8 & 49.6 & 46.3 & 59.7 & 54.7 & \textbf{57.1} & \_ & \_ & \_\\
\hline
\end{tabular}
\end{table}

\noindent\textbf{Drug-Gene-Mutation:}
Peng et al.~\cite{peng2017cross} released a dataset\footnote{\url{http://hanover.azurewebsites.net}} where known interactions among {\small\sf\em Drug}, {\small\sf\em Gene} and {\small\sf\em Mutation} were captured as a traditional mention-level ternary cross-sentence relation {\small\sf\em Interact}. 
We converted these annotations for the {\small\sf\em Interact} relation to {\em RIGD}-level. As we do not have access to complete documents, pseudo-documents were created for each section of consecutive sentences used in the dataset. As the original annotations were obtained through distant supervision; instead of using all non-positive candidate relation instances as negative instances, we only used the explicitly annotated negative relation instances. 
In order to compare the performance with the approach of Peng et al.~\cite{peng2017cross}, we computed the ``Accuracy'' metric at mention-level along with other metrics at {\em RIGD}-level (Table~\ref{tabDGM_results}). 

\noindent\textbf{MUC-6:}
We used the {\em train} and {\em test} datasets for management succession event from MUC-6~\cite{sundheim1996overview} and converted the annotations into a 4-ary relation {\small\sf\em Succession} (Table~\ref{example}). The event arguments in the original dataset are mapped to entity arguments for the relation. 32 documents each from train and test partitions contained the complete {\small\sf\em Succession} relation and were used for our evaluation (Table~\ref{tabMUC_results}). The original dataset labelled ``gold'' aliases for entities; these were used for converting the annotations. But our algorithm does not use gold aliases during testing
; rather we use simple high-precision domain-specific rules (e.g. {\small\tt Bert-Olof Svanholm} and {\small\tt Mr. Svanholm} are identified as aliases).
\begin{table}\small\center
\caption{\label{tabDGM_results}Mention-level accuracy \& RIGD-level (P,R,F) for the ternary {\small\sf\em Interact} relation (Drug-Gene-Mutation). The results by Quirk and Poon~\cite{quirk2017distant} are mentioned as reported by Peng el al.~\cite{peng2017cross}}
\begin{tabular}{lcccc}
\hline
{\bf Approach} & {\bf P} & {\bf R} & {\bf F} & {\bf Accuracy}\\
\hline
Maxent & 73.2 & 65.2 & 69.0 & 76.0 \\
LSTM & 72.9 & 67.4 & 70.0 & 77.5 \\
SVM with CSK & 75.5 & 67.1 & 71.1 & 78.8 \\
Quirk and Poon\cite{quirk2017distant} & - & - & - & 77.7 \\
Peng et al.\cite{peng2017cross} & - & - & - & 82.4 \\
Song et al.\cite{song2018n} & - & - & - & \textbf{83.2} \\
\hline
\end{tabular}
\normalsize
\end{table}
\begin{table}\small\center
\caption{\label{tabMUC_results}RIGD-level (P,R,F) for the 4-ary {\small\sf\em Succession} relation (MUC-6)}
\begin{tabular}{lcccc}
\hline
{\bf Approach} & {\bf P} & {\bf R} & {\bf F}\\
\hline
Maxent & 31.8 & 46.1 & 37.6\\
LSTM & 20.0 & 28.9 & 25.2\\
SVM with CSK & 73.3 & 28.9 & \textbf{41.5}\\
\hline
\end{tabular}
\normalsize
\end{table}


\noindent \textbf{Dataset Statistics} are shown in Table~\ref{tabDataStats}.

\noindent \textbf{Implementation Details:} 
Minimal spans used for filtering candidate relation instances were as follows (\#sentences): {\small\sf\em Succession}-$2$, {\small\sf\em Lives\_In}-$4$, {\small\sf\em Interact}-$2$. For all the experiments, the CSK parameters were set as follows: $N'=4$, $\lambda=0.9$. We used libsvm~\cite{CC01a} and keras~\cite{chollet2015keras} for SVM and LSTM implementations 
and our own implementation for MaxEnt. $C$ parameter for SVM was set to $1$ for all experiments and instance weights were used (in all classifiers) so as to ensure that sum of weights of positive instances is same as that of negative instances. 100-dim GloVe~\cite{pennington2014glove} word vectors (pre-trained on Wikipedia) 
were used for finding word clusters as well as for the LSTM-based classifier.

\subsection{Analysis of Results and Errors}
\noindent \textbf{Results:} The SVM with CSK outperformed MaxEnt and LSTM-based classifiers on all the 3 datasets (Tables~\ref{tabBB_dataset},~\ref{tabDGM_results} and~\ref{tabMUC_results}). For Bacteria-Biotope dataset, it outperformed the previous work, VERSE~\cite{lever2016verse}. For Drug-Gene-Mutation dataset, it outperformed one out of three previous works and for MUC-6, there is no comparable previous result.
Also, the SVM with CSK was observed to have higher precision than recall. This is because each possible subsequence (satisfying the constraints) of the tokens leads to a separate dimension in the transformed space using CSK. Thus, feature-space representation 
using CSK is sparse and it leads to higher precision and lower recall for SVM, with limited training data. 
There is scope for improving the recall by better generalizing the sequence representations.
\begin{table}\small\center
\caption{\label{tabDataStats}Dataset statistics. Except for the {\small\sf\em Interact} relation where negative instances are explicitly annotated, for other relations negative instances are automatically generated.}
\begin{tabular}{lccccc}
\hline
\multirow{2}{*}{\textbf{Relation}} & \multirow{2}{*}{\textbf{\#Documents}} & \multicolumn{2}{c}{\textbf{Mention-level Instances}} & \multicolumn{2}{c}{\textbf{RIGD-level Instances}}\\
\cline{3-6}
  &  & \#pos & \#neg & \#pos & \#neg \\
\hline
{\sf\em Lives\_In} (Bacteria-Biotope) & 107 & 814 & 2506 & 392 & 1517 \\
{\sf\em Interact} (Drug-Gene-Mutation) & 3652 & 3407 & 3564 & 2187 & 3182 \\
{\sf\em Succession},Train (MUC-6) & 36 & 377 & 43940 & 66 & 8344 \\
{\sf\em Succession},Test (MUC-6) & 36 & 474 & 13734 & 76 & 2874 \\
\hline
\end{tabular}
\normalsize
\end{table}

\noindent \textbf{Ablation Analysis:} 
We perform ablation analysis to evaluate contributions of key design elements: constraints and word clusters.
For all the datasets, the reported results (Tables~\ref{tabBB_dataset},~\ref{tabDGM_results} and~\ref{tabMUC_results}) are with word clusters and using the constraints.
Table~\ref{ablationResults} shows the effect of discarding each of these design elements. Constraints in CSK were observed to be beneficial for all the datasets, whereas word clusters were useful only for {\small\sf\em Lives\_In} relation. 
\begin{table}\small\center
\caption{\label{ablationResults}Ablation Analysis (all the numbers are at RIGD-level)}
\begin{tabular}{lccccccccc}
\hline
 & \multicolumn{3}{c}{{\small\sf\em Lives\_In}} & \multicolumn{3}{c}{{\small\sf\em Interact}} & \multicolumn{3}{c}{{\small\sf\em Succession}}\\
\textbf{Setting} & \multicolumn{3}{c}{(Bacteria-Biotope)} & \multicolumn{3}{c}{(Drug-Gene-Mutation)} & \multicolumn{3}{c}{(MUC-6)}\\
\cline{2-10}
  & \textbf{P} & \textbf{R} & \textbf{F} & \hspace{4mm}\textbf{P}\hspace{4mm} & \textbf{R} & \textbf{F} & \textbf{P} & \textbf{R} & \textbf{F}\\
\hline
\textbf{M: SVM with CSK} & 59.7 & 54.7 & 57.1 & 75.5 & 67.1 & 71.1 & 73.3 & 28.9 & 41.5 \\
M without Constraint 1 (SVM with GSK) & 74.6 & 12.7 & 21.6 & 73.1 & 57.3 & 64.2 & 100.0 & 7.9 & 14.6 \\
M without word clusters & 54.7 & 54.5 & 54.5 & 75.3 & 67.7 & 71.3 & 70.6 & 31.6 & 43.6 \\
\hline
\end{tabular}
\end{table}


\noindent \textbf{Error Analysis:} We analyzed poorer performance for the {\small\sf\em Succession} relation and observed that two major reasons are: {\em Class Imbalance} and presence of two arguments with the same entity type {\small\sf\em PER}. As it is a 4-ary relation, number of possible candidate relation instances is high and very few of them actually represent the relation; resulting in {\em Class Imbalance}. For $T_1$ (Table~\ref{seqRepresentation}), if last two arguments ($E_3$ and $E_4$) are swapped, we get almost identical sequence representation with just $E_3$ and $E_4$ swapping their positions. This new instance (with swapped $E_3$ and $E_4$) is a negative instance for the {\small\sf\em Succession} relation unlike $T_1$. It is challenging for the classifiers to distinguish between these nearly identical sequence representations of opposite classes, with limited training data.

We analyzed poorer performance for the {\small\sf\em Interact} relation as compared to the state-of-the-art and observed that a major reason was absence of ``gold'' entities information in the original dataset which only annotated entities which are part of annotated relation instances and not {\em all} entities. 
Tokens of the type $OE_{ET_j}$ in our sequence representation depend on information of mentions of all entities; hence the sequence representation could not characterize the relation instance completely. 
Also, the annotation labels obtained through distant supervision are not perfect.
E.g., we get a false positive for predicting $\langle${\tt\small ipilimumab}, {\tt\small BRAF}, {\tt\small V600E}$\rangle$ but it is a {\em true} relation instance as per the following sentence in the document: {\tt\small Rapid improvement of therapeutic responses using combined vemurafenib plus ipilimumab therapy for BRAF V600E mutation positive melanoma is expected.} 


Although, the LSTM-based classifier and the SVM with CSK both use the same sequence representation, the LSTM-based classifier performs poorly in comparison. Through the constraint on subsequences, the SVM with CSK harnesses the knowledge that the tokens of type $E_i$ are more important. Whereas the LSTM-based classifier does not explicitly harness this knowledge and needs further exploration.

\section{Related Work}
\label{sec:related_work}
Our formulation of the relation extraction task differs from the past work on the {\em distant supervision} based relation extraction~\cite{mintz2009distant,riedel2010modeling,hoffmann2011knowledge,surdeanu2012multi} and {\em slot filling} tasks~\cite{surdeanu2013overview} in terms of scope of the relations.  Rather than extracting corpus-level facts / relations, our approach focuses on determining whether a relation holds for a tuple of entities within scope of a particular document. Also, these approaches extract only binary relations which are expressed in single sentences. 
To the best of our knowledge, the problem of cross-sentence relation extraction was first addressed by Swampillai and Stevenson~\cite{swampillai2011extracting}. 
They proposed to introduce a dependency link between the root nodes of parse trees containing the given pair of entities. They developed features based on the shortest path connecting the pair of entities in the new ``fused'' tree. 
Recently, Quirk and Poon~\cite{quirk2017distant} proposed a new approach for cross-sentence relation extraction using distant supervision. They proposed a graph representation which incorporates both standard dependencies and discourse relations. 
In a document graph, each node is labeled with its lexical item, lemma and POS tag. Edges are added between adjacent words as well as between words connected with dependencies. 
Inter-sentential edges are added in 3 cases: i) edge between root nodes of adjacent sentences, ii) discourse relations and iii) co-references. Features are then extracted from multiple paths in this graph and a binary logistic regression classifier is trained using these features. 

Peng et al.~\cite{peng2017cross} proposed a general framework for N-ary cross-sentence relation extraction, based on 
graph LSTMs. 
They used the same document graph as proposed by Quirk and Poon~\cite{quirk2017distant} as a backbone for their graph LSTM. The word embeddings of input text are provided to the input layer. Next layer is formed by the graph LSTM which learns a contextual representation for each word. For the entities in a relation instance, their contextual representations are concatenated and provided as the input to the relation classifier. 
For training graph LSTMs using backpropagation, the document graph needs to be partitioned into 2 directed acyclic graphs (DAGs). Song et al.~\cite{song2018n} proposed graph-state LSTMs which do not need such partitioning and use the original graph. They used a parallel state to model each word, where state values are enriched recurrently via message passing.
In contrast to these mention-level N-ary and cross-sentence relation extraction approaches, our proposed formulation captures a broader view of any relation instance in a single representation; by incorporating multiple mentions of the entity arguments and their aliases in a document.

\section{Conclusion and Future Work}
We proposed a new formulation of the relation extraction task, 
where the relations are more general than intra-sentence relations in the sense that they may span multiple sentences (\textbf{cross-sentence}) and may have more than two arguments (\textbf{N-ary}). Also, these relations are more specific than corpus-level relations because their scope is only limited within a document. 
A novel approach as well as new schemes for annotation and evaluation were proposed for this proposed formulation. 
We designed a sequence representation for characterizing instances of such relations and explored various classifiers whose features are derived from this sequence representation. We also designed the Constrained Subsequence Kernel for the SVM classifier. We evaluated our approach on three datasets across two domains. 
In future, we plan to explore various directions identified in the results analysis section.

%
%
%
\bibliographystyle{splncs04}
\bibliography{ref}

\section*{Appendix}
\subsection*{Alias Detection Rules}
We use a few high-precision rules for identifying aliases. The information of aliases is needed at several stages of our approach: 

\noindent$\bullet$ Identifying positive / negative instances for training the classifiers: Alias information is needed to check whether any candidate relation instance is similar to any relation instance listed in gold-annotations (i.e. in $LR_i$). Only when it is not similar to any of the relation instances in $LR_i$, it is treated as a negative instance for classifiers.

\noindent$\bullet$ Construction of sequence representations for candidate relation instances: Alias information is needed for adding tokens of the type $E_i$ and $OE_{ET_j}$ at appropriate positions in sequence representations, as described in Section 3.2.

\noindent$\bullet$ Evaluation: Alias information is also needed for evaluation as described in Section 2.


Following are the details of the high-precision rules used for identifying aliases.

\noindent \textbf{Bacteria Biotope dataset:} Two entity mentions are aliases of each other if:
\begin{itemize}
\item The two entity mentions are exactly same
\item If the first word of one entity mention is a short-form of the first word of another entity mention and rest all the words in these mentions are same. Here, short-form is just the first character followed by ``.'' (e.g. {\small\tt Salmonella Typhimurium} $\leftrightarrow$ {\small\tt S. Typhimurium})
\item One entity mention is prefix of another and its length is more than half of the length of the other entity mention
\end{itemize}
The above rules are used while constructing sequence representations as well as while evaluation. 

\noindent \textbf{Drug-Gene-Mutation dataset:} Two entity mentions are aliases of each other only if one of the entity mentions is prefix of another. This rule is used while constructing sequence representations as well as while evaluation. 

\noindent \textbf{MUC-6 dataset:} Two entity mentions are aliases of each other if:
\begin{itemize}
\item The two entity mentions are exactly same
\item Any one is the prefix of the other
\item Special case: one starts with {\small\tt Chief Executive} and another is {\small\tt CEO}
\item If one entity mention has the first word as {\small\tt Mr.} or {\small\tt Ms.} and its last word is the suffix of another mention
\item Except for the case where both the entity mentions are of type {\small\sf\em POST}: if one entity mention is suffix of another
\end{itemize}
The above rules are used while evaluation. In addition to the above rules, we also use Stanford CoreNLP coreferences for construction of sequence representations, i.e. for adding tokens of the type $E_i$ and $OE_{ET_j}$ at appropriate positions in sequence representations.
\end{document}